\documentclass[final]{cvpr}

\usepackage{times}
\usepackage{epsfig}
\usepackage{graphicx}
\usepackage{amsmath}
\usepackage{amssymb}

\usepackage{graphicx}
\usepackage{amsmath}
\usepackage{amsthm}
\usepackage{amsfonts} 
\usepackage{mathabx}
\usepackage{bm}
\usepackage{subcaption}
\usepackage{color}
\usepackage{multirow}
\usepackage{stackengine}
\usepackage[switch]{lineno}

\usepackage{makecell}

\newcommand\blfootnote[1]{%
  \begingroup
  \renewcommand\thefootnote{}\footnote{#1}%
  \addtocounter{footnote}{-1}%
  \endgroup
}

\usepackage[pagebackref=true,breaklinks=true,colorlinks,bookmarks=false]{hyperref}

\def\cvprPaperID{6370} 
\def\confYear{CVPR 2021}

\begin{document}

\title{Towards Extremely Compact RNNs for Video Recognition with Fully Decomposed Hierarchical Tucker Structure}

\author{Miao Yin$^1$, Siyu Liao$^{2\dagger}$, Xiao-Yang Liu$^3$, Xiaodong Wang$^3$ and Bo Yuan$^1$\\
$^1$Rutgers University, $^2$Amazon, $^3$Columbia University\\
{\tt\small miao.yin@rutgers.edu, liasiyu@amazon.com, \{xl2427, xw2008\}@columbia.edu,}\\ {\tt\small bo.yuan@soe.rutgers.edu}
}

\maketitle
\pagestyle{empty}  
\thispagestyle{empty} 

\begin{abstract}
Recurrent Neural Networks (RNNs) have been widely used in sequence analysis and modeling. However, when processing high-dimensional data, RNNs typically require very large model sizes, thereby bringing a series of deployment challenges. Although  various prior works have been proposed to reduce the RNN model sizes, executing RNN models in the resource-restricted environments is still a very challenging problem. In this paper, we propose to develop extremely compact RNN models with \textit{fully decomposed hierarchical Tucker (FDHT) structure.} The HT decomposition does not only provide much higher storage cost reduction than the other tensor decomposition approaches, but also brings better accuracy performance improvement for the compact RNN models. Meanwhile, unlike the existing tensor decomposition-based methods that can only decompose the input-to-hidden layer of RNNs, our proposed fully decomposition approach enables the comprehensive compression for the entire RNN models with maintaining very high accuracy. Our experimental results on several popular video recognition datasets show that, our proposed fully decomposed hierarchical tucker-based LSTM (FDHT-LSTM) is extremely compact and highly efficient. To the best of our knowledge, FDHT-LSTM, for the first time, consistently achieves very high accuracy with only few thousand parameters (3,132 to 8,808) on different datasets. Compared with the state-of-the-art compressed RNN models, such as TT-LSTM, TR-LSTM and BT-LSTM, our FDHT-LSTM simultaneously enjoys both order-of-magnitude (3,985$\times$ to 10,711$\times$) fewer parameters and significant accuracy improvement (0.6\% to 12.7\%). \blfootnote{$^\dagger$This work was done when the author was with Rutgers University.}
\end{abstract}


\section{Introduction}
\label{sec:intro}

Recurrent Neural Networks (RNNs), especially their advanced
variants such as Long-Short Term Memory (LSTM)
and Gated Recurrent Unit (GRU), have achieved unprecedented
success in sequence analysis and processing. Thanks
to their powerful capability of capturing and modeling the
temporary dependency and correlation in the sequential data,
the state-of-the-art RNNs have been widely deployed in many
important artificial intelligence (AI) fields, such as natural language processing (NLP) \cite{sutskever2011generating}, speech recognition \cite{mikolov2011extensions}, and computer vision \cite{yu2016video}.

Despite their current prosperity, the efficient deployment of RNNs is still facing several challenges, especially the \textit{large model size problem}. Due to the widespread existence of high-dimensional input data in many applications, e.g. NLP and video processing, the input-to-hidden weight matrices of RNNs are often extremely large. For instance, as pointed out in \cite{yang2017tensor}, even with small-size hidden layer such as 256 hidden states, an LSTM working on UCF11 video recognition dataset \cite{liu2009recognizing} already requires more than 50 million parameters. Such ultra-high model size, consequently, brings a series of deployment challenges for RNNs, including but not limited to high difficulty of training, susceptibility to overfitting, long processing latency and inefficient energy consumption etc.

In order to address this problem, several prior works \cite{donahue2015long}, \cite{yue2015beyond} use convolutional neural network (CNNs) as the front-end feature extractors to reduce the size of input data of back-end RNNs. Although this strategy indeed reduces the model size as compared to the pure RNN-based solution, the back-end RNN part is still very large. Leveraging the popular model compression approaches in CNNs, such as pruning and quantization, is another alternative. However, the compression ratio provided by these methods is still insufficient,  considering the very large size of the original RNN models.

\textbf{Tensor Decomposition-based RNNs.} Due to the above described limitations, recent RNN compression studies have mainly focused on a new direction -- building the compact RNN models using low-rank \textit{tensor decomposition}. By its nature, tensor decomposition can represent a very large tensor with the combination of multiple very small tensor cores. Correspondingly, the number of the required representation parameters can be significantly reduced. Based on this unique and powerful capability, various compact RNN models have been developed using different tensor decomposition approaches. In \cite{yang2017tensor}, TT-LSTM and TT-GRU, which decompose input-to-hidden layer of RNNs to Tensor Train (TT) format, are proposed for video recognition. Similarly, BT-LSTM \cite{ye2018learning} and TR-LSTM \cite{ye2018learning} are also developed by decomposing the input-to-hidden layer via Block-Term Tensor (BT) and Tensor Ring (TR) formats, respectively. Besides those compact RNN models for computer vision tasks, in \cite{yu2017long} a TT-structure LSTM is developed for long-term forecasting in dynamic systems. Compared with the original large-size RNNs, these compact models show significant model size reduction with maintaining competitive classification/prediction accuracy.

\begin{figure}[t]
    \centering
    \includegraphics[width=\linewidth]{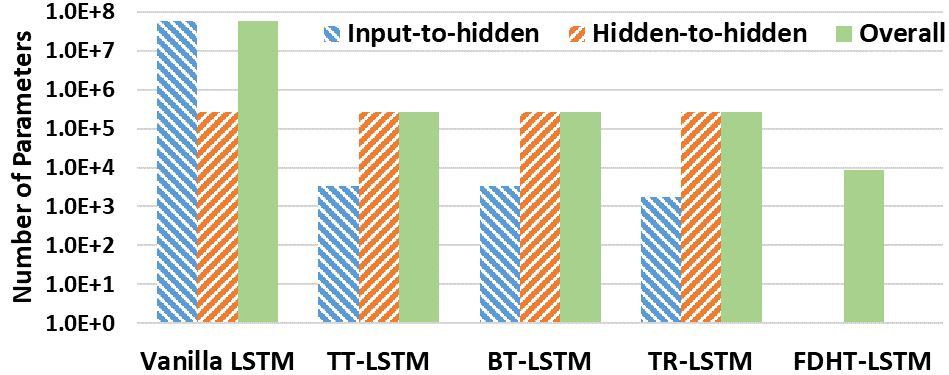}
    \vspace{-6mm}
    \caption{With compressing only input-to-hidden layer, hidden-to-hidden layer becomes the bottleneck of the state-of-the-art compact RNN models. The details of vanilla and compressed LSTM models are described in Table \ref{tab:direct_ucf11}.}
    \label{fig:partial_compress}
    \vspace{-3mm}
\end{figure}

\textbf{Limitations of Prior Works.} Despite their promising potentials, the state-of-the-art tensor decomposition-based RNN models are still facing two inherent limitations: (1) Only the input-to-hidden layers, instead of the entire RNNs, are decomposed to small tensor cores. Consequently, there are still a large amount of parameters in the uncompressed hidden-to-hidden layers of RNNs, thereby making the large model size problem may still exist, especially in the resource-constrained scenarios (see Figure \ref{fig:partial_compress}). Performing additional tensor decomposition on hidden-to-hidden layers is an alternative solution; however, as will be shown in the next section, straightforward decomposing on both input-to-hidden and hidden-to-hidden layers causes significant accuracy degradation. (2) The underlying tensor decomposition approaches used in the state-of-the-art compact RNN models have inherent constraints and limitations. For instance, TT decomposition requires the border tensor cores have to be rank-1, thereby directly hindering the representation power of TT-LSTM. Also, BT-LSTM models suffers computation overhead due to the extra flatten and permutation operations incurred by BT structure. More generally, from the perspective of tensor theory, neither TT, TR nor BT decomposition provides the best space complexity reduction. Consequently, the existing tensor decomposition-based solutions are still not ideal for designing high-accuracy ultra-compact RNN models.

\begin{figure}[t]
    \centering
    \includegraphics[width=\linewidth]{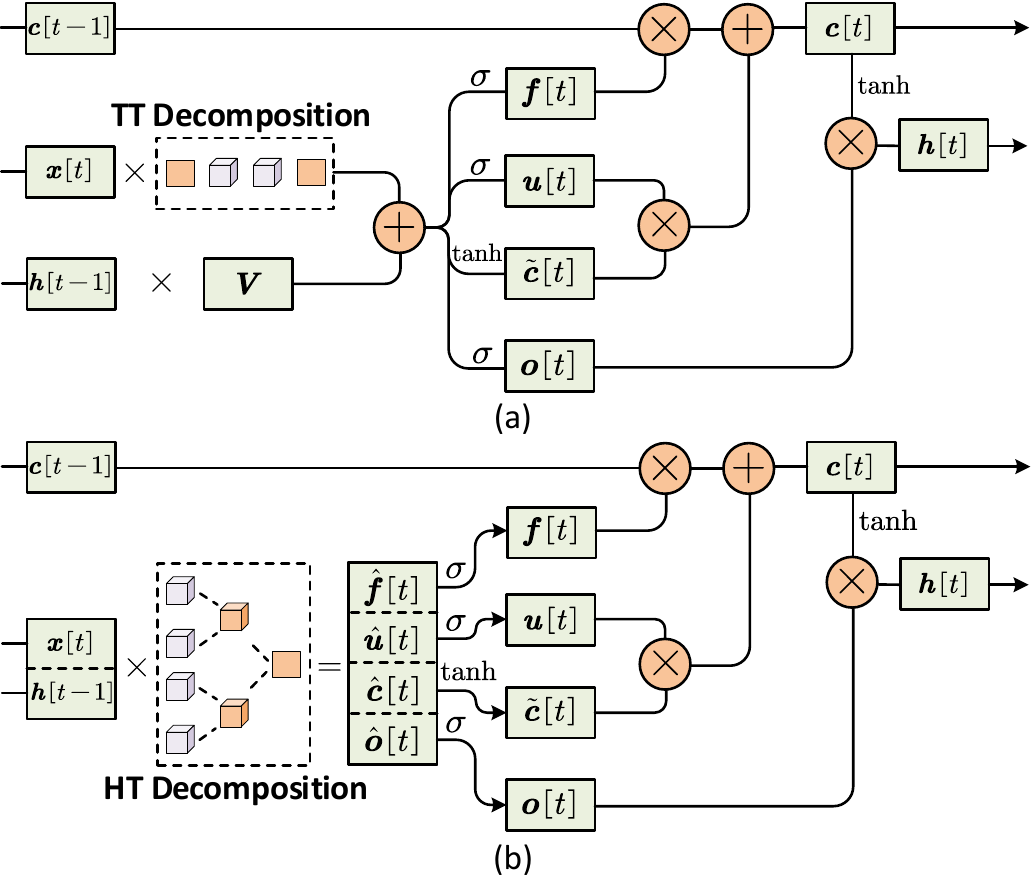}
    \vspace{-7mm}
    \caption{Architecture of tensor decomposition-based LSTM. $\bm{f}$ is the forget gate's activation vector, $\bm{u}$ is the input gate's activation vector, $\tilde{\bm{c}}$ is the cell input activation vector, and $\bm{o}$ is the output gate's activation vector in LSTM. (a) State-of-the-art TT-LSTM. (b) The proposed fully decomposed hieararchical Tucker LSTM (FDHT-LSTM). }
    \label{fig:cht_lstm}
    \vspace{-3mm}
\end{figure}

\textbf{Technical Preview \& Benefits.} To overcome these limitations, in this paper we propose to develop extremely compact RNN models with \textit{fully decomposed hierarchical Tucker (FDHT) structure}. As shown in Figure \ref{fig:cht_lstm}, our proposed FDHT-structure RNN models have two main features. \underline{First}, \textit{Hierarchical Tucker} (HT) decomposition \cite{hackbusch2009new}, a little explored but powerful tool for capturing and modeling the correlation and structure in high-dimensional data, is used to build the underlying RNN structure. \underline{Second}, the entire RNN models, instead of one or few component layers, are constructed in the HT structure in a homogeneous way. Such exploration on the low-rank correlation among multiple component layers of RNN models is non-trivial, and brings order-of-magnitude parameter reduction over the state-of-the-art approaches with still maintaining very high accuracy. The benefits of the proposed FDHT-based RNN models are summarized as follows:

\begin{itemize}
   
     \item \textbf{Benefits of Hierarchical Tucker Structure.} The underlying HT structure enables the RNN models enjoy much fewer model parameters and simultaneously higher accuracy than the state of the art. Compared with its well-explored counterparts (e.g. TT, TR and BT decomposition) adopted in the prior works, HT decomposition inherently provides higher space complexity reduction on the same-size tensor data with the same selected rank. By leveraging such theoretical advantage, the large-size RNN models can be constructed with very few parameters in the HT format. Additionally, the inherent hierarchical structure of HT also enables better weight sharing and hierarchical representation from high-dimensional data, thereby significantly improving RNN models' representation capability. As verified by our empirical experiments on different datasets, the LSTM models built on HT structure consistently outperform the existing LSTM models using other tensor decomposition in terms of both classification accuracy and model size reduction.

     \item \textbf{Benefits of Full Decomposition.} Built on the top of the underlying HT structure, enabling full decomposition with maintaining high accuracy brings further performance improvement for our proposed compact RNN models. As analyzed before, the existing tensor decomposition-based RNN models are limited by their uncompressed hidden-to-hidden layers. Although the same tensor decomposition can be performed on each individual uncompressed layers to further reduce model size, as will be shown in the experiment, such layer-wise decomposition causes significant accuracy degradation. Unlike this straightforward strategy, our proposed full decomposition approach integrates different RNN layers together, and then compresses the entire RNN model in a homogeneous way. Such integration-based full decomposition maintains the homogeneity of the model and avoid the accuracy loss. Consequently, our fully decomposed HT-structure RNN models further require even much fewer parameters than the non-fully decomposed HT-structure model with still remaining high accuracy.

\end{itemize}

\textbf{Realizing Few Thousand Parameters-only RNN Models.} By jointly using these two approaches, we develop extremely compact RNN models with high accuracy. Experiments show that, our proposed FDHT-LSTM consistently achieves very high accuracy with only very few parameters (3,132 to 8,808) on different video recognition datasets. To the best of our knowledge, it is the first RNN model that can only use few thousand parameters to achieve high accuracy on video recognition tasks. Compared with
the state-of-the-art compressed RNN models, such as TT-LSTM, TR-LSTM and BT-LSTM, our FDHT-LSTM simultaneously enjoys both order-of-magnitude (3,985$\times$ to 10,711$\times$) fewer parameters and significant accuracy improvement (0.6\% to 12.7\%).

\textbf{Difference from Other Tucker/HT RNN/CNN Works.} Recently in learning theory community, HT decomposition has been used to analyze the expressive power of RNNs and CNNs \cite{nadav2018analysis}, \cite{cohen2016expressive} and \cite{khrulkov2017expressive}. However, these prior works focus on representing an \textbf{uncompressed} neural network models with HT format to explore the theoretical property such as expressive power. Instead, our work focus on building \textbf{compressed} RNN models via performing HT decomposition to its layers. Also, using Tucker decomposition to compress CNN/RNN is studied in  \cite{tjandra2018tensor}, \cite{kim2015compression}. However, these works utilized Tucker decomposition, instead of Hierarchical Tucker decomposition, to compress models. In tensor theory, HT decomposition is different from Tucker decomposition, and HT enjoys much higher space complexity reduction than Tucker decomposition. Therefore, the compression ratio of our FDHT-RNN is much higher than the prior Tucker decomposition-based models.

\section{Compact Fully Decomposed Hierarchical Tucker RNN Model}
\label{sec:HTRNN}

\subsection{Preliminaries}
\label{subsec:prelim}

\indent \textbf{Notation.} Throughout the paper we use boldface calligraphic script letters, boldface capital letters, and boldface lower-case letters to represent tensors, matrices, and vectors, respectively, e.g.  $\bm{\mathcal{X}}\in\mathbb{R}^{n_1 \times n_2 \times \cdots \times n_d}$,  $\bm{X}\in\mathbb{R}^{n_1\times n_2}$, and $\bm{x}\in\mathbb{R}^{n_1}$. Also,  $\bm{\mathcal{X}}_{(i_1,\cdots,i_d)}\in\mathbb{R}$ denotes the entry of tensor  $\bm{\mathcal{X}}$. 

Similarly, $\bm{X}_{(i,j)}$ represents the entry of matrix $\bm{X}$.

 \textbf{Tensor Contraction.} An HT-decomposed tensor is essentially the consecutive product of multiple tensor contraction results, where tensor contraction is executed between two tensors with at least one matched dimension. For instance, given two tensors $\bm{\mathcal{A}}\in\mathbb{R}^{n_1\times n_2 \times l}$ and $\bm{\mathcal{B}}\in\mathbb{R}^{l\times m_1\times m_2}$, where the 3rd dimension of $\bm{\mathcal{A}}$ matches the 1st dimension of $\bm{\mathcal{B}}$ with length $l$, the tensor contraction result is a size- $n_1\times n_{2}\times m_1 \times m_2$ tensor as
$$ 
(\bm{\mathcal{A}}\times_{1}^{3}\bm{\mathcal{B}})_{(i_1, i_2, j_1, j_2)}=\sum_{\alpha=1}^{l}\bm{\mathcal{A}}_{(i_1, i_2, \alpha)}\bm{\mathcal{B}}_{(\alpha, j_1, j_2)}.
$$

\begin{figure}[t!]
    \centering
    \includegraphics[width=\linewidth]{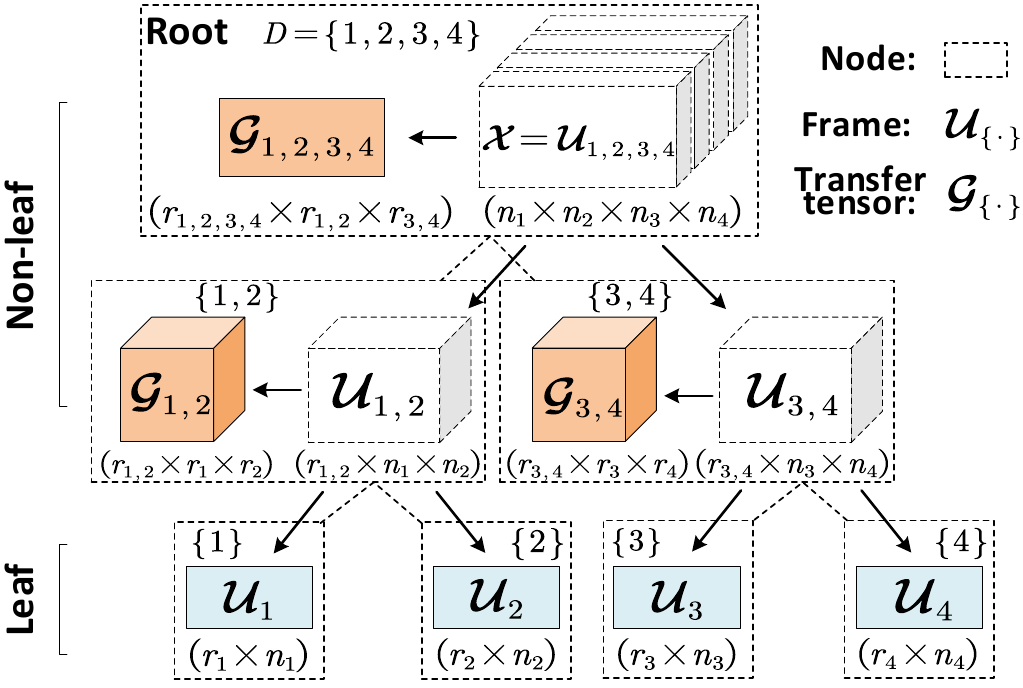}
    \vspace{-5mm}
    \caption{Example HT decomposition with a tensor of 4 orders. All the dashed lines and boxes describe a binary tree with root $D=\{1, 2, 3, 4\}$, where the dashed boxes represent the nodes. Here node $\{1\}$ is a leaf node, whose parent and sibling are node $\{1,2\}$ and node $\{2\}$, respectively. Here $\bm{\mathcal{X}}$ is decomposed to a set of orange-colored transfer tensors and blue-colored leaf frames. }
    \label{fig:htd_example}
    \vspace{-2mm}
\end{figure}

\textbf{Hierarchical Tucker Decomposition.} The Hierarchical Tucker decomposition is a special type of tensor decomposition approach with hierarchical levels with respect to the order of the tensor. As illustrated in Figure \ref{fig:htd_example}, an HT-decomposed tensor can be recursively decomposed into intermediate components, referred as \textit{frames}, from top to bottom in a binary tree, where each frame corresponds to a unique \textit{node}, and each node is associated with a \textit{dimension set}. 
In general, for a HT-decomposed tensor $\bm{\mathcal{X}}\in\mathbb{R}^{n_1\times\cdots\times n_d}$, we can build a binary tree with a root node associated with $D=\{1,2,\cdots,d\}$ and $\bm{\mathcal{X}}=\bm{\mathcal{U}}_D$ as the root frame. We define $s\subsetneq D$ is associated with the node corresponding to $\bm{\mathcal{U}}_s$, and $s_1, s_2\subsetneq s$ are associated with the left and right child nodes of the $s$-associated node. Hence, as $\mu_s=\min(s), \nu_s=\max(s)$, 
each non-leaf frame $\bm{\mathcal{U}}_{s}\in\mathbb{R}^{r_{s}\times n_{\mu_s}\times\cdots\times n_{\nu_s}}$ can be recursively decomposed to its left and right child frames ($\bm{\mathcal{U}}_{s_1}$ and $\bm{\mathcal{U}}_{s_2}$) and transfer tensor $\bm{\mathcal{G}}_s\in\mathbb{R}^{r_s\times r_{s_1} \times r_{s_2}}$ as

\begin{equation}
\bm{\mathcal{U}}_s=\bm{\mathcal{G}}_s\times_{1}^{2}\bm{\mathcal{U}}_{s_1}\times_{1}^{2}\bm{\mathcal{U}}_{s_2}.
\label{eqn:htd}
\end{equation}

Consequently, by performing this recursive decomposition till the bottom of the binary tree, we can decompose the original ${n_1\times\cdots\times n_d}$-order tensor $\bm{\mathcal{X}}=\bm{\mathcal{U}}_D$ into the combination of the 2-order leaf frames and 3-order transfer tensors. Notice that here $r_s$, as \textit{hierarchical rank}, is an important parameter that determines the decomposition effect.

\subsection{Compact HT-structure Linear Layer}

In this subsection we describe the details of building a compact HT-structure linear layer, which will be used as the foundation to build the compact FDHT-RNN models.

\textbf{Tensorization.} In general, the key idea of building HT structured linear layer is to transform the weight matrix $\bm{W}\in\mathbb{R}^{M\times N}$ to the HT-based format. Considering $\bm{W}$ is a 2-D matrix, while HT decomposition is mainly performed on high-order tensor, we first need to reshape $\bm{W}$ as well as its affiliated input vector $\bm{x}\in\mathbb{R}^{N}$ and output vector $\bm{y}\in\mathbb{R}^{M}$ to tensor format as $\bm{\mathcal{W}}\in\mathbb{R}^{m_1\times\cdots\times m_d\times n_1\times\cdots\times n_d}$,  $\bm{\mathcal{X}}\in\mathbb{R}^{n_1\times\cdots\times n_d}$ and $\bm{\mathcal{Y}}\in\mathbb{R}^{m_1\times\cdots\times m_d}$, respectively, where $M=\prod_{i=1}^{d}m_i$ and $N=\prod_{j=1}^{d}n_j$.


\textbf{Decomposing  $\bm{W}$.} Given a tensorized $\bm{W}$ as $\bm{\mathcal{W}}\in\mathbb{R}^{m_1\times\cdots\times m_d\times n_1\times\cdots\times n_d}$, we can now leverage HT decomposition to represent the large-size $\bm{W}$ using a set of small-size matrices and tensors. In general, following Equation (\ref{eqn:htd}), $\bm{\mathcal{W}}$ can be decomposed as
\begin{equation}
\begin{aligned}
\bm{\mathcal{W}}_{(i_1,\cdots,i_d,j_1,\cdots,j_d)}=
\sum_{k=1}^{r_D}\sum_{p=1}^{r_{D_1}}\sum_{q=1}^{r_{D_2}}(\bm{\mathcal{G}}_D)_{(k,p,q)}\\
\cdot(\bm{\mathcal{U}}_{D_1})_{(p,\varphi_{D_1}(\bm{i},\bm{j}))}(\bm{\mathcal{U}}_{D_2})_{(q,\varphi_{D_2}(\bm{i},\bm{j}))},
\label{eqn:w_ht}
\end{aligned}
\end{equation}
\noindent where $\varphi_s(\bm{i},\bm{j})$ is a mapping function that produces the correct indices $\bm{i}=(i_1,\cdots,i_d)$ and $\bm{j}=(j_1,\cdots, j_d)$ for a specified frame $\bm{\mathcal{U}}_s$ with the given $s$ and $d$. For instance, with $d=6$ and $s=\{3,4\}$, the output of $\varphi_s(\bm{i},\bm{j})$ is $(i_3,i_4,j_3,j_4)$. In addition, $\bm{\mathcal{U}}_{D_1}$ and $\bm{\mathcal{U}}_{D_2}$ can be recursively computed as
\begin{equation}
\begin{aligned}
(\bm{\mathcal{U}}_s)_{(k,\varphi_s(\bm{i},\bm{j}))}
=\sum_{p=1}^{r_{s_1}}\sum_{q=1}^{r_{s_2}}(\bm{\mathcal{G}}_{s})_{(k,p,q)}\\
\cdot(\bm{\mathcal{U}}_{s_1})_{(p,\varphi_{s_1}(\bm{i},\bm{j}))}(\bm{\mathcal{U}}_{s_2})_{(q,\varphi_{s_1}(\bm{i},\bm{j}))},
\end{aligned}
\end{equation}
\noindent where $D=\{1,2,\cdots,d\}$, $D_1=\{1,\cdots,\lfloor d/2\rfloor\}$ and $D_2=\{\lceil d/2\rceil,\cdots,d\}$ are associated with left and right child nodes of the root node.


\textbf{HT-structure Layer.} With the HT-decomposed weight matrix, a linear layer with HT structure can now be developed. Specifically, the HT-format matrix-vector multiplication, as the kernel computation in the forward propagation procedure on the layer, is performed as follows:
\begin{equation}
\begin{aligned}
\bm{\mathcal{Y}}_{(\bm{i})}=\sum_{\bm{j}}
\sum_{k=1}^{r_D}\sum_{p=1}^{r_{D_1}}\sum_{q=1}^{r_{D_2}}(\bm{\mathcal{G}}_D)_{(k,p,q)}\\
\cdot(\bm{\mathcal{U}}_{D_1})_{(p,\varphi_{D_1}(\bm{i},\bm{j}))}(\bm{\mathcal{U}}_{D_2})_{(q,\varphi_{D_2}(\bm{i},\bm{j}))}\bm{\mathcal{X}}_{(\bm{j})}.
\end{aligned}
\end{equation}

Considering the desired output $\bm{y}$ of HT-structure layer is a vector, the calculated $\bm{\mathcal{Y}}$ needs to be re-shaped again to the 1-D format. Consequently, we denote the entire forward computing procedure from input $\bm{x}$ to output $\bm{y}$ as
\begin{equation}
\bm{y}= HTL(\bm{W}, \bm{x}).
\end{equation}

\begin{figure}[t!]
    \centering
    \includegraphics[width=\linewidth]{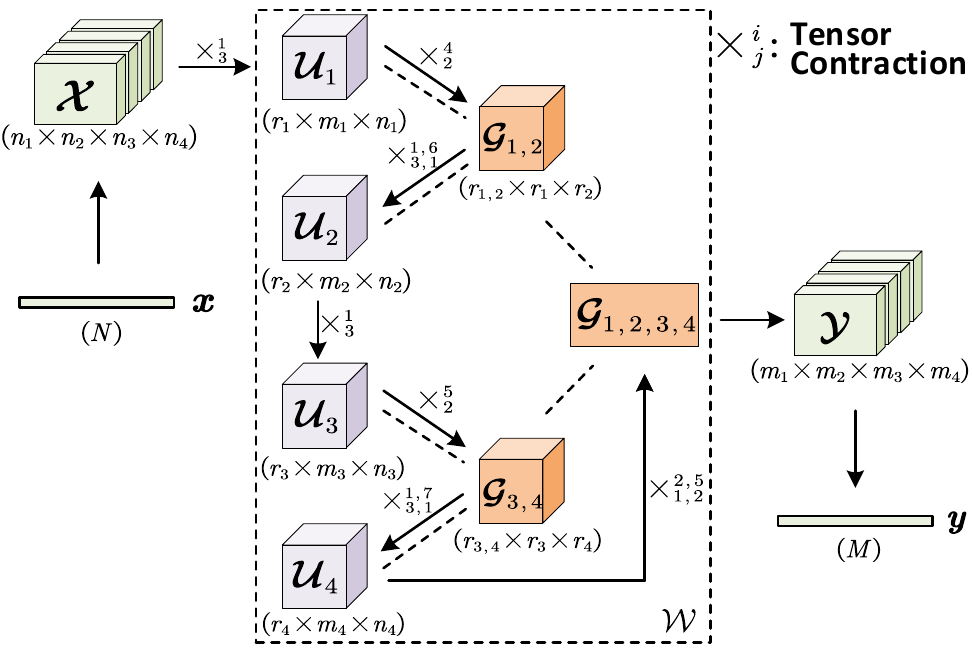}
    \caption{Example computation process in an HT-structure layer with $d$=4. Solid lines with arrow connect two tensors that are contracted. Note that different from Figure \ref{fig:htd_example}, the leaf frames $\bm{\mathcal{U}}$ here is 3-dimensional. This is because the dimensions of input tensor $\bm{\mathcal{X}}$ here is different.}
    \label{fig:htl}
\end{figure}

Figure \ref{fig:htl} illustrates the computation process in an HT-structure layer. Here the arrows in a sequence represent the whole computation flow. As shown in this figure, the input vector is first tensorized, and then tensor contraction is performed in a hierarchical way. Finally the tensorized output is obtained.

\begin{figure}[t]

    \centering
    \includegraphics[width=\linewidth]{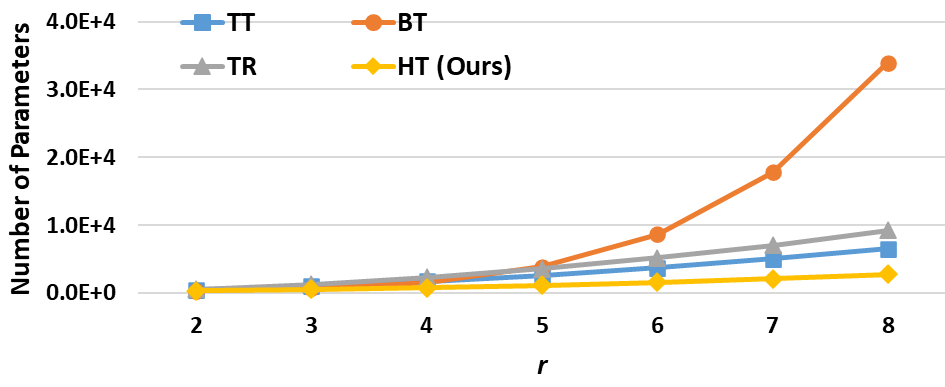}
    \vspace{-4mm}
    \caption{Comparison on number of parameters with different tensor decomposition for the same weight matrix. All the tensor decomposition methods use the same setting $d=5$, $(n_1,\cdots,n_5)=(8,10,10,9,8)$, $(m_1,\cdots,m_5)=(4,4,2,4,2)$, and $r$ is the rank.}
    \label{fig:complexity}
    \vspace{-1.5mm}
\end{figure}

\textbf{Benefits on Low Cost.} One major benefit of using HT-structure linear layer is that its inherent high complexity reduction. As shown in Table \ref{tab:complexity}, compared with the vanilla uncompressed linear layer as well as other tensor decomposition-based layers, using HT structure can bring the lower space complexity with the same rank setting. Besides theoretical complexity analysis, we also verify this low-cost benefits of HT structure via empirical experiments. Figure \ref{fig:complexity} shows the number of parameters to store a compact weight matrix. Here we adopt the size-$57,600 \times256$ weight matrix used in \cite{yang2017tensor} \cite{ye2018learning} \cite{pan2019compressing} for evaluation. From Figure \ref{fig:complexity} it is seen that HT-based approach indeed require the fewest parameters than other tensor decomposition-based methods. Our evaluation results in the Experiments Section also verify such advantages on compression ratios over various datasets.

\begin{table}
\setlength\tabcolsep{4.6pt}
\def\arraystretch{1.15}
\centering
\begin{tabular}{|c|c|} 
\hline

\textbf{Compressed Linear Layer}  & \multicolumn{1}{c|}{\textbf{Space Complexity}}  \\
\hline\hline
Uncompressed &  \multirow{1}{*}{$\mathcal{O}(NM)$} \\
\hline
Tensor Train (TT)-structure &  \multirow{1}{*}{$\mathcal{O}(dmnr^2)$} \\
\hline
Tensor Ring (TR)-structure &  \multirow{1}{*}{$\mathcal{O}(dmnr^2)$} \\
\hline
Block-Term (BT)-structure &  \multirow{1}{*}{$\mathcal{O}(dmnr+r^d)$} \\
\hline
Hierarchical Tucker (HT)-structure & \multirow{1}{*}{$\mathcal{O}(dmnr+dr^3)$} \\

\hline
\end{tabular}
\vspace{-2mm}
\caption{Comparison of space complexity among different tensor decomposition-based linear layers. $C$ is the CP rank-value defined in BT decomposition, and $r=\max_{s\subsetneq D}r_s$, $m=\max_{k\in D}m_{k}$, $n=\max_{k\in D}n_{k}$.}
\vspace{-2.5mm}
\label{tab:complexity}

\end{table}

\subsection{Fully Decomposing the HT-structure RNN}

\begin{figure}[t]
    \centering
    \includegraphics[width=\linewidth]{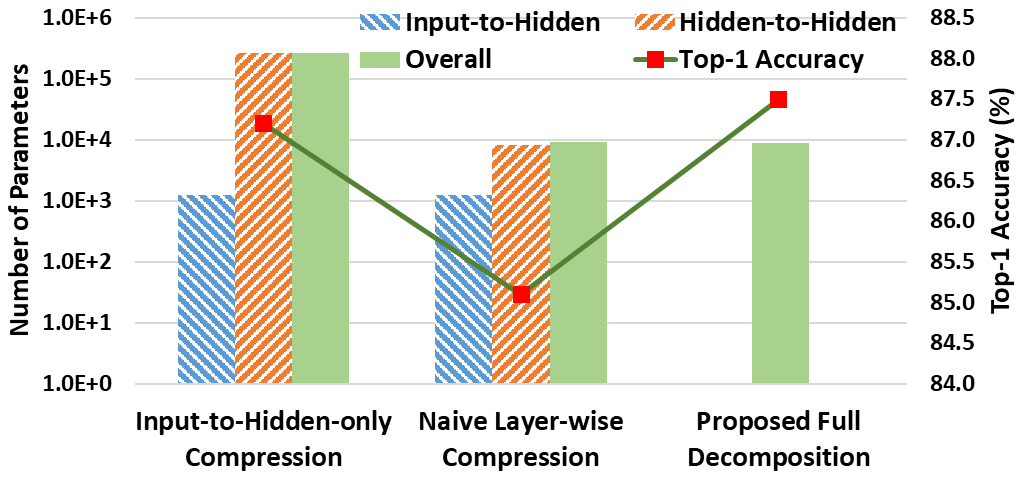}
    \caption{Comparison of number of parameters and accuracy between layer-wise compression and the proposed full decomposition. Evaluation dataset is UCF11.}
    \label{fig:full_compress}
    \vspace{-2.5mm}
\end{figure}

\textbf{Challenges of Fully Compressing RNN.} Based on the proposed HT-structure layer, next we aim to compress the entire RNN models using HT decomposition. A straightforward way is to simply build each component layer of RNNs with HT format. In other words, all of the weight matrices of input-to-hidden and hidden-to-hidden layers are decomposed into HT structure.  Although indeed providing further compression on the hidden-to-hidden layers, such layer-wise compression strategy suffers huge accuracy drop. As shown in Figure \ref{fig:full_compress}, the layer-wise compression using HT decomposition causes $2.4\%$ accuracy drop as compared to the input-to-hidden-only compression. Therefore, realizing the full compression of the entire RNN models without accuracy drop is non-trivial but challenging.

\textbf{Proposed FDHT-RNN.} To achieve that, we propose to develop the entire compact RNNs in a homogeneous way, namely \textit{fully decomposed HT (FDHT)} structure. Take LSTM, as the most popular and advanced variant of RNNs, for example. As illustrated in Figure \ref{fig:cht_lstm}, our key idea is to first concatenate all the weight matrix as a single big matrix. In other words, the entire LSTM model can be interpreted as a single "mega" linear layer. From the perspective of forward propagation, at each time step, all the intermediate results can be viewed as being calculated using only one matrix multiplication:
\begin{equation}
\begin{aligned}
\begin{bmatrix}
\hat{\bm{f}}[t]\\
\hat{\bm{u}}[t]\\
\hat{\bm{c}}[t]\\
\hat{\bm{o}}[t]
\end{bmatrix}&=
\begin{bmatrix}
\bm{W}_f & \bm{V}_f\\
\bm{W}_u & \bm{V}_u\\
\bm{W}_c & \bm{V}_c\\
\bm{W}_o & \bm{V}_o
\end{bmatrix}
\begin{bmatrix}
\bm{x}[t]\\
\bm{h}[t-1]
\end{bmatrix}\\
&=\bm{W}\bm{I}[t].
\end{aligned}
\label{eqn:entirety}
\end{equation}

Based on the above interpretation, the HT structure can be imposed on this integrated layer. Specifically, we can tensorize and decompose the entire RNN models to HT format and perform forward propagation as follows:
\begin{equation}
    \begin{bmatrix}
\hat{\bm{f}}[t]\\
\hat{\bm{u}}[t]\\
\hat{\bm{c}}[t]\\
\hat{\bm{o}}[t]
\end{bmatrix} = \bm{Z}[t]=HTL(\bm{W}, \bm{I}[t]).
\end{equation}
\noindent After this HT-based computation, the outputs of the FDHT-LSTM can be calculated as follows:
\begin{equation}
\begin{aligned}
\bm{f}[t]=&\sigma(\hat{\bm{f}}[t]+\bm{b}_f)\\
\bm{u}[t]=&\sigma(\hat{\bm{u}}[t]+\bm{b}_u)\\
\bm{c}[t]=&\bm{f}[t]\odot\bm{c}[t-1]+\bm{u}[t]\odot\tanh(\hat{\bm{c}}[t]+\bm{b}_c)\\
\bm{o}[t]=&\sigma(\hat{\bm{o}}[t]+\bm{b}_o)\\
\bm{h}[t]=&\bm{o}[t]\odot\tanh(\bm{c}[t]),
\end{aligned}
\end{equation}
\noindent where $\sigma$, $\tanh$ and $\odot$ are the sigmoid function, hyperbolic function and element-wise product, respectively.

\underline{HT-based Gradient Calculation.}
To ensure the valid training on FDHT-RNN, the gradient calculation in the backward propagation should also be accordingly reformulated to HT-based format. In general, considering for HT-structure linear layer $\bm{\mathcal{W}}=\bm{\mathcal{U}}_D$ and $\frac{\partial\bm{\mathcal{Y}}}{\partial\bm{\mathcal{U}}_D}=\bm{\mathcal{X}}$, assuming $s$ is associated with a left node, as we denote that $F(s)$ and $B(s)$ are the sets associated with the parent and sibling nodes of the $s$-associated node in the binary tree, respectively, and define  $\mu_s=\min(s), \nu_s=\max(s)$, the partial derivative of output tensor with respect to frames can be calculated in the following recursive way until $F(s)$ is equal to $D$:
\begin{equation}
\begin{aligned}
\frac{\partial\bm{\mathcal{Y}}}{\partial\bm{\mathcal{U}}_s}=&
\bm{\mathcal{G}}_{F(s)}\times_{1}^{3}\bm{\mathcal{U}}_{B(s)}\\
&\times_{\substack{1,\cdots,\nu_{B(s)}-\mu_{B(s)}+2,\nu_{F(s)}-\mu_{F(s)}+3,\\\cdots,\nu_{F(s)}-\mu_{F(s)}+\nu_{B(s)}-\mu_{B(s)}+3}}^{1,3,\cdots,2\nu_{B(s)}-2\mu_{B(s)}+4}\frac{\partial\bm{\mathcal{Y}}}{\partial\bm{\mathcal{U}}_{F(s)}}.
\end{aligned}
\label{eqn:rec_derivative}
\end{equation}
\noindent Based on Equation (\ref{eqn:rec_derivative}), the gradients for leaf frames and transfer tensors can be computed as follows: 
\begin{align}
\frac{\partial L}{\partial\bm{\mathcal{U}}_s}=&\frac{\partial\bm{\mathcal{Y}}}{\partial\bm{\mathcal{U}}_s}\times_{1,\cdots,\mu_s-1,\nu_s+1,\cdots,d}^{\nu_s-\mu_s+3,\cdots,d+1}\frac{\partial L}{\partial\bm{\mathcal{Y}}},\label{eqn:gradients_u}\\
\frac{\partial L}{\partial\bm{\mathcal{G}}_s}=&\frac{\bm{\mathcal{Y}}}{\partial\bm{\mathcal{U}}_s}\times_{2,\cdots,\nu_{s_1}-\mu_{s_1}+2}^{2,\cdots,\nu_{s_1}-\mu_{s_1}+2} \bm{\mathcal{U}}_{s_1}\nonumber\\
&\times_{2,\cdots,\nu_{s_2}-\mu_{s_2}+2}^{3,\cdots,\nu_{s_2}-\mu_{s_2}+3}\bm{\mathcal{U}}_{s_2}\times_{1,\cdots,d}^{4,\cdots,d+3}\frac{\partial L}{\partial \bm{\mathcal{Y}}}.\label{eqn:gradients_g}
\end{align}

In general, our proposed "integrate-then-decompose" strategy brings significant performance benefit for the compact FDHT-RNN models. As illustrated in Figure \ref{fig:full_compress}, unlike layer-wise compression that has significant accuracy drop, our proposed full decomposition approach does not bring any performance loss. Instead, it even outperforms the input-to-hidden-only compression counterpart with much fewer parameters. As will be shown in the next section, evaluation on various datasets also show that our full decomposed HT-format RNN models achieve the same or even higher accuracy than the model that has only HT structure on the input-to-hidden layers. We hypothesize such phenomenon may result from the special architecture of FDHT-RNN: when the entire RNN is constructed in the HT structure, the multi-dimensional low-rank correlations across the entire model can be explored and captured in a more precise and comprehensive way, thereby enabling very compact RNN model without affecting accuracy performance.


\section{Experiments}
\label{sec:exp}

In this section, we evaluate the performance of FDHT-RNN on different datasets, and compare them with the state-of-the-art with respect to compression ratio and test accuracy. Considering LSTM is the current most commonly used RNN variant in both academia and industry, our experiments focus on FDHT-LSTM, and compare it with the vanilla uncompressed LSTM and recent advances in compressed RNN such as TT-LSTM \cite{yang2017tensor}, BT-LSTM \cite{ye2018learning} and TR-LSTM \cite{pan2019compressing}. Also, similar to prior works, we also train and evaluate a model that only has HT structure in the input-to-hidden layer, namely HT-LSTM, for ablation study.

\textbf{Selection of Tensor Parameters $d$ and $r$.} In our experiments we set $d$, as the dimension number of the tensors that the input, output and weight matrix are tensorized to, as the same to the setting in \cite{yang2017tensor} and \cite{ye2018learning}. Also similar to prior works, we select $r$ via empirical setting to make good balance between compression ratio and model accuracy.



\textbf{Training Strategy.} Following the similar setting in prior works, we adopt two types of training strategy: end-to-end direct training and training with pre-trained CNN. In the end-to-end direct training the input of LSTM is the raw data, e.g. video clips; while training with pre-trained CNNs means the back-end LSTM receives the compact features extracted by a front-end pre-trained CNN.

\subsection{End-to-End Direct Training (RNN-only Model)}

\textbf{Hyperparameter Setting.} We train the models using ADAM optimizer with L2 regularization of coefficient 0.001. Also, dropout rate is set as 0.25 and batch size is 16.

\textbf{UCF11 Dataset.} The UCF11 dataset \cite{liu2009recognizing} consists of 11-class human action (e.g. biking, diving, basketball) videos with totally 1,600 video clips. Each class is assembled by 25 video groups, where each group contains at least 4 action clips with resolution as $320\times 240$. 

At data pre-processing stage we choose the same settings used in the related work \cite{ye2018learning} \cite{pan2019compressing} for fair comparison. Specifically, the resolution of video clips is first scaled down to $160 \times 120$, and then 6 frames from each clip are randomly sampled to form the sequential input.

For the baseline vanilla uncompressed LSTM model, it contains 4 input-to-hidden layers and 4 hidden-to-hidden layers, where the size of its input vector is $160\times 120\times 3=57,600$, and the number of hidden states in each layer is 256. For our proposed FDHT-LSTM, in order to factorize the size of concatenated vector $\bm{I}$, we pad the input vector to size $16 \times 16 \times 16 \times 15 - 256 = 61184$ by zeros. The concatenated vector is reshaped to a tensor of shape $16\times 16\times 16\times 15$, and the hidden state is reshaped to a tensor of shape $4\times 4\times 4\times 4$. All leaf and non-leaf ranks are set as 14 and 12, respectively.

\begin{table}[t]
\centering
\def\arraystretch{1.15}
\begin{tabular}{|c|c|c|c|}
\hline
\multirow{2}{*}{\textbf{Model}}  & \multicolumn{2}{c|}{\textbf{Number of Parameters}}   & \textbf{Top-1} \\
\cline{2-3}
 & \textbf{Input-Hidden} & \textbf{Overall} & \textbf{Acc.} (\%) \\
\hline
\hline
LSTM  & 58.98M     & 59.24M  & 69.7     \\
\hline
\multirow{2}{*}{TT-LSTM}  & \multirow{2}{*}{3,360} & 265.50K & \multirow{2}{*}{79.6}    \\
& & (223$\times$) &\\
\hline
\multirow{2}{*}{BT-LSTM}  & \multirow{2}{*}{3,387}  & 265.53K  & \multirow{2}{*}{85.3}     \\
 & & (223$\times$) &\\
\hline
\multirow{2}{*}{TR-LSTM}  & \multirow{2}{*}{1,725}  & 263.87K & \multirow{2}{*}{86.9}     \\
 & & (225$\times$) & \\
\hline
HT-LSTM  & \multirow{2}{*}{1,245}  & 263.39K & \multirow{2}{*}{87.2}     \\
(Ours) & & (225$\times$) & \\
\hline
\hspace{-1mm}FDHT-LSTM\hspace{-1mm}   & \multirow{2}{*}{-} & \textbf{8,808} & \multirow{2}{*}{\textbf{87.5}}    \\
(Ours) & & (\textbf{6,726$\times$}) & \\
\hline
\end{tabular}
\vspace{-2mm}
\caption{
Performance of different RNN compression works on UCF11 dataset using end-to-end direct training.}
\label{tab:direct_ucf11}
\vspace{-2.5mm}
\end{table}

Table \ref{tab:direct_ucf11} summarizes the performance of our FDHT-LSTM on UCF11 dataset and compare it with the related works. It is seen that compared with vanilla LSTM using 59 million parameters, FDHT-LSTM only needs 8,808 parameters with 17.8\% accuracy increase. Compared with the recent advances on compressing RNNs using other tensor decomposition methods, including TT-LSTM, BT-LSTM and TR-LSTM, our proposed FDHT-LSTM requires at least 6,501$\times$ fewer parameters with at least 0.6\% accuracy increase.

\textbf{Youtube Celebrities Face Dataset.} Youtube dataset \cite{kim2008face} contains 1,910 video clips from 47 subjects. Also, the resolutions of the frames vary for different video clips. Being consistent with prior works, for data pre-processing the resolution of the input data to FDHT-LSTM is re-scaled as $160\times 120$. Also, 6 frames in each video clips are randomly sampled to form the input sequence. 

In this experiment we build a FDHT-LSTM with the similar setting with the one used in UCF11 dataset. A slight difference is here all non-leaf ranks are set as 11. Table \ref{tab:directly_ytc} shows the performance comparison with TT-LSTM \cite{yang2017tensor} on this dataset. From this table it is seen that FDHT-LSTM achieves 7,117$\times$ compression ratio over the original uncompressed LSTM with much higher accuracy. Compared with TT-LSTM, FDHT-LSTM has 6,894$\times$ fewer parameters with 12.7\% higher accuracy. 

\begin{table}[t]
\def\arraystretch{1.15}
\centering
\begin{tabular}{|c|c|c|c|}  
\hline
\multirow{2}{*}{\textbf{Model}}  & \multicolumn{2}{c|}{\textbf{Number of Parameters}}   & \textbf{Top-1} \\
\cline{2-3}
 & \textbf{Input-Hidden} & \textbf{Overall} & \textbf{Acc.} (\%) \\
\hline
\hline
LSTM &  58.98M  & 59.24M  & 33.2       \\
\hline
\multirow{2}{*}{TT-LSTM} & \multirow{2}{*}{3,392}  & 265.54K  & \multirow{2}{*}{75.5}      \\
& & (223$\times$) & \\
\hline
HT-LSTM & \multirow{2}{*}{810}  & 262.95K  & \multirow{2}{*}{88.1}      \\
(Ours) & & (225$\times$) & \\
\hline
\hspace{-1mm}FDHT-LSTM\hspace{-1mm}  & \multirow{2}{*}{-}  & \textbf{8,324}  & \multirow{2}{*}{\textbf{88.2}}      \\
(Ours) & & (\textbf{7,117$\times$}) & \\
\hline
\end{tabular}
\vspace{-2mm}
\caption{
Performance of different RNN compression work on Youtube celebrities face dataset using end-to-end direct training. \textbf{Notice that no prior works report performance of BT-LSTM and TR-LSTM on this dataset.}}
\label{tab:directly_ytc}

\end{table}

Besides, on the same Youtube dataset we also compare FDHT-LSTM with several other reported works without using tensor decomposition method. As shown in Table \ref{tab:directly_ytc2}, among those works the state-of-the-art model is \cite{li2018face}, which has the highest reported accuracy (84.6\%). Compared with that model, FDHT-LSTM achieves 3.6\% higher test accuracy with using much fewer parameters.

\begin{table}[t]
\def\arraystretch{1.15}
\centering
\begin{tabular}{|l|c|c|}  
\hline
 \multicolumn{1}{|c|}{\multirow{2}{*}{\textbf{Model}}}  & \multirow{2}{*}{\makecell{\textbf{Number of}\\\textbf{Parameters}}}& \textbf{Top-1} \\
 & & \textbf{Acc.} (\%)\\
\hline
\hline
DML-PV  & 220K &  82.8 \\
\hline
VGGFACE + RRNN  & $\geq$42M  & 84.6 \\
\hline
VGG16-GCR  & 138M & 82.9  \\
\hline
HT-LSTM (Ours) & 263K & 88.1 \\
\hline
FDHT-LSTM (Ours)  & \textbf{8,324} & \textbf{88.2} \\
\hline
\end{tabular}
\vspace{-2mm}
\caption{Comparison between HT-LSTM using end-to-end direct training and other models without using tensor decomposition on Youtube celebrities face dataset, such as DML-PV\cite{cheng2017duplex}, VGGFACE + RRNN \cite{li2018face} and VGG16-GCR \cite{liu2019group}.}
\label{tab:directly_ytc2}
\vspace{-2.5mm}
\end{table}

\subsection{Training with Pre-trained CNNs (CNN+RNN)}
Another set of our experiments is based on training strategy using pre-trained CNNs as the front-end feature extractor. As indicated in \cite{donahue2015long}, using the front-end CNN can reduce the required input vector size of RNN and  significantly improve the overall performance of the entire CNN+RNN model.

\textbf{Hyperparameter Setting.} In this part of experiments dropout rate is set as 0.5. The L2 regularization with coefficient 0.0001 is used, and the entire FDHT-LSTM model is trained using ADAM optimizer with batch size 16.

\textbf{UCF11 Dataset.} Consistent with \cite{pan2019compressing}, Inception-V3 \cite{szegedy2016rethinking} is selected as the the front-end CNN, whose output is a flattened size-2,048 feature vector. For the baseline  vanilla uncompressed  LSTM  model, the size of hidden state is set as 2048. For our proposed FDHT-LSTM, the concatenated vector is of size $2048 + 2048 = 4096$, and is reshaped to a tensor of size $8\times 8\times 8\times 8$. Similarly, the output vector is reshaped to a tensor of size $4\times 8\times 8\times 8$. In addition, all leaf ranks are set as 9, and all non-leaf ranks are set as 6.

Table \ref{tab:cnn_ucf11} summarizes the test accuracy of different models over UCF11 dataset. It is seen that with pre-trained CNN model as front-end feature extractor, FDHT-LSTM achieves 98.4\% accuracy, which is 3.8\% higher than the best reported accuracy from the state of the art. Compared with the TR-LSTM using the same front-end CNN, FDHT-LSTM achieves 4.6\% accuracy increase. Meanwhile, as shown in Table \ref{tab:performance_mix}, for such CNN + RNN model, TR-LSTM only brings 1.9$\times$ fewer parameters, while FDHT achieves 2.5$\times$ parameter reduction.

\begin{table}[t]
\def\arraystretch{1.15}
\centering
\begin{tabular}{|l|c|c|}  
\hline
 \multicolumn{1}{|c|}{\multirow{2}{*}{\textbf{Model}}}  & \multirow{2}{*}{\makecell{\textbf{Number of}\\\textbf{Parameters}}}& \textbf{Top-1} \\
 & & \textbf{Acc.} (\%)\\
 \hline
 \hline
Soft Attention  & 311M  &  85.0   \\
\hline
Deep Fusion  &  179M  &    94.6  \\
\hline

CNN + LSTM & 55M & 92.3 \\
\hline
CNN + TR-LSTM & 39M &  93.8 \\
\hline
CNN + HT-LSTM (Ours)& 22M & 98.1 \\
\hline
CNN + FDHT-LSTM (Ours)& \textbf{22M}  &  \textbf{98.4} \\

\hline
\end{tabular}
\vspace{-2mm}
\caption{Comparison between FDHT-LSTM using front-end pre-trained CNN and other related works on UCF11 dataset, such as Soft Attention \cite{sharma2015action}, Deep Fusion \cite{gammulle2017two} and CNN + LSTM/TR-LSTM \cite{pan2019compressing}. \textbf{Notice that no prior works report performance of CNN + TT-LSTM and CNN + BT-LSTM on this dataset.}}
\label{tab:cnn_ucf11}
\vspace{-2.5mm}
\end{table}

\textbf{HMDB51 Dataset.} HMDB51 dataset \cite{kuehne2011hmdb} contains 6,849 video clips that belong to 51 action categories, where each of them consists of more than 101 clips. Again, for the experiment on this dataset we use Inception-V3 as the pre-trained CNN model, whose extracted feature is flattened to a length-2,048 vector. Reshaped from concatenated vector, the tensor has size of $8\times 8\times 8\times 8$. We also reshape the output vector to a tensor of size $4\times 8\times 8\times 8$. Meanwhile, all leaf ranks are set as 14, and all non-leaf ranks are set as 12.

Table \ref{tab:cnn_hmdb51} summarizes the performance of CNN-aided FDHT-LSTM and other related works on this dataset. It is seen that FDHT-LSTM achieves 64.2\% accuracy, which obtains 0.4\% increase than the recent TR-LSTM. Note that two-stream I3D is a 3D-CNN model which is more advanced, while our model only uses 2D pre-trained CNN. With fewer parameters, our approach can still achieve the comparable performance. As shown in Table \ref{tab:performance_mix}, for such CNN+RNN model, TR-LSTM only brings 1.4$\times$ fewer parameters, while FDHT-LSTM achieves 2.5$\times$ parameter reduction.


\begin{table}[h]
\def\arraystretch{1.15}
\centering
\begin{tabular}{|l|c|c|}  
\hline
 \multicolumn{1}{|c|}{\multirow{2}{*}{\textbf{Model}}}  & \multirow{2}{*}{\makecell{\textbf{Number of}\\\textbf{Parameters}}}& \textbf{Top-1} \\
 & & \textbf{Acc.} (\%)\\
\hline
\hline
TDD + FV  & 117M &  63.2   \\
\hline
VGG + Two-Stream Fusion &181M & 62.1 \\
\hline
Two-Stream I3D & 25M & \textbf{66.4} \\
\hline
CNN + LSTM & 55M& 62.9 \\
\hline
CNN + TR-LSTM & 39M &  63.8 \\
\hline
CNN + HT-LSTM (Ours) & 22M &  64.2 \\
\hline
CNN + FDHT-LSTM (Ours) & \textbf{22M} &  64.2 \\
\hline
\end{tabular}
\vspace{-2mm}
\caption{Comparison among HT-LSTM using front-end pre-trained CNN and other related works on HMDB51 dataset, such as TDD + FV \cite{wang2015action}, VGG + Two-Stream Fusion \cite{feichtenhofer2016convolutional}, Two-Stream I3D \cite{carreira2017quo} and CNN + LSTM/TR-LSTM \cite{pan2019compressing}. \textbf{Notice that no prior works report performance of CNN + TT-LSTM and CNN + BT-LSTM on this dataset.}}
\label{tab:cnn_hmdb51}

\end{table}

\begin{table}[h]
\def\arraystretch{1.15}
\centering

\begin{tabular}{|c|c|c|c|c|}
\hline
\multirow{3}{*}{\makecell{\textbf{Model}\\(with CNN)}}  & \multicolumn{4}{c|}{\textbf{Number of Parameters}}  \\
\cline{2-5}
 & \multirow{2}{*}{\textbf{CNN}} &  \multicolumn{2}{c|}{\textbf{RNN}} & \multirow{2}{*}{\textbf{Total}}\\
 \cline{3-4}
 & &\multirow{1}{*}{\hspace{-2mm}\textbf{Input-Hidden}\hspace{-2mm}}& \textbf{Overall}  & \\
\hline
\hline
LSTM & \multirow{7}{*}{\hspace{-2mm}21.77M\hspace{-2mm}} & 16.78M &33.55M&\multirow{1}{*}{\hspace{-2mm}55.32M\hspace{-2mm}}\\
\cline{1-1}\cline{3-5}
\multirow{2}{*}{TR-LSTM} & & \multirow{2}{*}{671.2K} & 17.45M &\multirow{1}{*}{\hspace{-2mm}39.22M\hspace{-2mm}}\\
& & & (1.9$\times$) &\multirow{1}{*}{\hspace{-0.5mm}(1.4$\times$)\hspace{-0.5mm}}\\
\cline{1-1}\cline{3-5}
\multirow{1}{*}{\hspace{-2mm}FDHT-LSTM\hspace{-2mm}}& & \multirow{2}{*}{-} & \textbf{3,132} &\multirow{1}{*}{\hspace{-2mm}\textbf{21.77M}\hspace{-2mm}}\\
(UCF11)& & &\multirow{1}{*}{\hspace{-2mm}(\textbf{10,713$\times$})\hspace{-2mm}}&\multirow{1}{*}{\hspace{-0.5mm}(\textbf{2.5$\times$})\hspace{-0.5mm}}\\
\cline{1-1}\cline{3-5}
\multirow{1}{*}{\hspace{-2mm}FDHT-LSTM\hspace{-2mm}}& & \multirow{2}{*}{-} & \textbf{8,416} &\multirow{1}{*}{\hspace{-2mm}\textbf{21.78M}\hspace{-2mm}}\\
(HMDB51)& & &\multirow{1}{*}{\hspace{-2mm}(\textbf{3,987$\times$})\hspace{-2mm}}&\multirow{1}{*}{\hspace{-0.5mm}(\textbf{2.5$\times$})\hspace{-0.5mm}} \\
\hline

\end{tabular}
\vspace{-2mm}
\caption{Comparison of detailed model size among our FDHT-LSTM, TR-LSTM and vanilla LSTM with pre-trained CNN. The same models of CNN + LSTM and CNN + TR-LSTM are used for both UCF11 and HMDB51 datasets.}
\vspace{-2.5mm}
\label{tab:performance_mix}
\end{table}

\section{Conclusion}

In this paper, we propose a new and extremely compact RNN model with fully decomposed and Hierarchical Tucker structure. Our experiments on different datasets show that, our proposed FDHT-LSTM models significantly outperform the state-of-the-art compressed RNN models in terms of both compression ratio and test accuracy. To the best of our knowledge, FDHT-LSTM is the first RNN model that achieves very high accuracy with only few thousand parameters on different video recognition datasets. 

\section*{Acknowledgements}
This work was partially supported by National Science Foundation under Grant CCF-1854742 and SHF-7995357.

\clearpage
{\small
\bibliographystyle{ieee_fullname}
\bibliography{cvpr}

\begin{thebibliography}{10}\itemsep=-1pt

\bibitem{carreira2017quo}
Joao Carreira and Andrew Zisserman.
\newblock Quo vadis, action recognition? a new model and the kinetics dataset.
\newblock In {\em Proceedings of the IEEE Conference on Computer Vision and
  Pattern Recognition}, pages 6299--6308, 2017.

\bibitem{cheng2017duplex}
Gong Cheng, Peicheng Zhou, and Junwei Han.
\newblock Duplex metric learning for image set classification.
\newblock {\em IEEE Transactions on Image Processing}, 27(1):281--292, 2017.

\bibitem{cohen2016expressive}
Nadav Cohen, Or Sharir, and Amnon Shashua.
\newblock On the expressive power of deep learning: A tensor analysis.
\newblock In {\em Conference on Learning Theory}, pages 698--728, 2016.

\bibitem{donahue2015long}
Jeffrey Donahue, Lisa Anne~Hendricks, Sergio Guadarrama, Marcus Rohrbach,
  Subhashini Venugopalan, Kate Saenko, and Trevor Darrell.
\newblock Long-term recurrent convolutional networks for visual recognition and
  description.
\newblock In {\em Proceedings of the IEEE Conference on Computer Vision and
  Pattern Recognition}, pages 2625--2634, 2015.

\bibitem{feichtenhofer2016convolutional}
Christoph Feichtenhofer, Axel Pinz, and Andrew Zisserman.
\newblock Convolutional two-stream network fusion for video action recognition.
\newblock In {\em Proceedings of the IEEE Conference on Computer Vision and
  Pattern Recognition}, pages 1933--1941, 2016.

\bibitem{gammulle2017two}
Harshala Gammulle, Simon Denman, Sridha Sridharan, and Clinton Fookes.
\newblock Two stream lstm: A deep fusion framework for human action
  recognition.
\newblock In {\em IEEE Winter Conference on Applications of Computer Vision},
  pages 177--186. IEEE, 2017.

\bibitem{hackbusch2009new}
Wolfgang Hackbusch and Stefan K{\"u}hn.
\newblock A new scheme for the tensor representation.
\newblock {\em Journal of Fourier Analysis and Applications}, 15(5):706--722,
  2009.

\bibitem{khrulkov2017expressive}
Valentin Khrulkov, Alexander Novikov, and Ivan Oseledets.
\newblock Expressive power of recurrent neural networks.
\newblock {\em arXiv preprint arXiv:1711.00811}, 2017.

\bibitem{kim2008face}
Minyoung Kim, Sanjiv Kumar, Vladimir Pavlovic, and Henry Rowley.
\newblock Face tracking and recognition with visual constraints in real-world
  videos.
\newblock In {\em Proceedings of the IEEE Conference on Computer Vision and
  Pattern Recognition}, pages 1--8. IEEE, 2008.

\bibitem{kim2015compression}
Yong-Deok Kim, Eunhyeok Park, Sungjoo Yoo, Taelim Choi, Lu Yang, and Dongjun
  Shin.
\newblock Compression of deep convolutional neural networks for fast and low
  power mobile applications.
\newblock {\em arXiv preprint arXiv:1511.06530}, 2015.

\bibitem{kuehne2011hmdb}
Hildegard Kuehne, Hueihan Jhuang, Est{\'\i}baliz Garrote, Tomaso Poggio, and
  Thomas Serre.
\newblock Hmdb: a large video database for human motion recognition.
\newblock In {\em International Conference on Computer Vision}, pages
  2556--2563. IEEE, 2011.

\bibitem{li2018face}
Yang Li, Wenming Zheng, Zhen Cui, and Tong Zhang.
\newblock Face recognition based on recurrent regression neural network.
\newblock {\em Neurocomputing}, 297:50--58, 2018.

\bibitem{liu2019group}
Bo Liu, Liping Jing, Jia Li, Jian Yu, Alex Gittens, and Michael~W Mahoney.
\newblock Group collaborative representation for image set classification.
\newblock {\em International Journal of Computer Vision}, 127(2):181--206,
  2019.

\bibitem{liu2009recognizing}
Jingen Liu, Jiebo Luo, and Mubarak Shah.
\newblock Recognizing realistic actions from videos “in the wild”.
\newblock In {\em Proceedings of the IEEE Conference on Computer Vision and
  Pattern Recognition}, pages 1996--2003. IEEE, 2009.

\bibitem{mikolov2011extensions}
Tom{\'a}{\v{s}} Mikolov, Stefan Kombrink, Luk{\'a}{\v{s}} Burget, Jan
  {\v{C}}ernock{\`y}, and Sanjeev Khudanpur.
\newblock Extensions of recurrent neural network language model.
\newblock In {\em IEEE International Conference on Acoustics, Speech and Signal
  Processing}, pages 5528--5531. IEEE, 2011.

\bibitem{nadav2018analysis}
Cohen Nadav, Sharir Or, Levine Yoav, Tamari Ronen, Yakira David, and Shashua
  Amnon.
\newblock Analysis and design of convolutional networks via hierarchical tensor
  decompositions.
\newblock {\em arXiv preprint arXiv:1705.02302}, 2018.

\bibitem{pan2019compressing}
Yu Pan, Jing Xu, Maolin Wang, Jinmian Ye, Fei Wang, Kun Bai, and Zenglin Xu.
\newblock Compressing recurrent neural networks with tensor ring for action
  recognition.
\newblock In {\em Proceedings of the AAAI Conference on Artificial
  Intelligence}, volume~33, pages 4683--4690, 2019.

\bibitem{sharma2015action}
Shikhar Sharma, Ryan Kiros, and Ruslan Salakhutdinov.
\newblock Action recognition using visual attention.
\newblock In {\em Neural Information Processing Systems: Time Series Workshop},
  2015.

\bibitem{sutskever2011generating}
Ilya Sutskever, James Martens, and Geoffrey~E Hinton.
\newblock Generating text with recurrent neural networks.
\newblock In {\em International Conference on Machine Learning}, pages
  1017--1024, 2011.

\bibitem{szegedy2016rethinking}
Christian Szegedy, Vincent Vanhoucke, Sergey Ioffe, Jon Shlens, and Zbigniew
  Wojna.
\newblock Rethinking the inception architecture for computer vision.
\newblock In {\em Proceedings of the IEEE Conference on Computer Vision and
  Pattern Recognition}, pages 2818--2826, 2016.

\bibitem{tjandra2018tensor}
Andros Tjandra, Sakriani Sakti, and Satoshi Nakamura.
\newblock Tensor decomposition for compressing recurrent neural network.
\newblock In {\em 2018 International Joint Conference on Neural Networks
  (IJCNN)}, pages 1--8. IEEE, 2018.

\bibitem{wang2015action}
Limin Wang, Yu Qiao, and Xiaoou Tang.
\newblock Action recognition with trajectory-pooled deep-convolutional
  descriptors.
\newblock In {\em Proceedings of the IEEE Conference on Computer Vision and
  Pattern Recognition}, pages 4305--4314, 2015.

\bibitem{yang2017tensor}
Yinchong Yang, Denis Krompass, and Volker Tresp.
\newblock Tensor-train recurrent neural networks for video classification.
\newblock In {\em International Conference on Machine Learning}, pages
  3891--3900. JMLR. org, 2017.

\bibitem{ye2018learning}
Jinmian Ye, Linnan Wang, Guangxi Li, Di Chen, Shandian Zhe, Xinqi Chu, and
  Zenglin Xu.
\newblock Learning compact recurrent neural networks with block-term tensor
  decomposition.
\newblock In {\em Proceedings of the IEEE Conference on Computer Vision and
  Pattern Recognition}, pages 9378--9387, 2018.

\bibitem{yu2016video}
Haonan Yu, Jiang Wang, Zhiheng Huang, Yi Yang, and Wei Xu.
\newblock Video paragraph captioning using hierarchical recurrent neural
  networks.
\newblock In {\em Proceedings of the IEEE Conference on Computer Vision and
  Pattern Recognition}, pages 4584--4593, 2016.

\bibitem{yu2017long}
Rose Yu, Stephan Zheng, Anima Anandkumar, and Yisong Yue.
\newblock Long-term forecasting using higher order tensor rnns.
\newblock {\em arXiv preprint arXiv:1711.00073}, 2017.

\bibitem{yue2015beyond}
Joe Yue-Hei~Ng, Matthew Hausknecht, Sudheendra Vijayanarasimhan, Oriol Vinyals,
  Rajat Monga, and George Toderici.
\newblock Beyond short snippets: Deep networks for video classification.
\newblock In {\em Proceedings of the IEEE Conference on Computer Vision and
  Pattern Recognition}, pages 4694--4702, 2015.

\end{thebibliography}
}

\end{document}